\documentclass{article}

\usepackage[utf8]{inputenc} %
\usepackage[T1]{fontenc}    %
\usepackage{hyperref}       %
\usepackage{url}            %
\usepackage{booktabs}       %
\usepackage{amsfonts}       %
\usepackage{nicefrac}       %
\usepackage{microtype}      %
\usepackage{xcolor}         

\usepackage{amssymb}
\usepackage{amsmath}
\usepackage{amssymb}
\usepackage{amsmath}
\usepackage{booktabs}
\usepackage{algorithm}
\usepackage{algorithmic}
\usepackage{mathtools}
\usepackage{multirow}
\usepackage{amsthm}

\theoremstyle{definition}
\newtheorem{theorem}{Theorem}[section]

\newtheorem{definition}[theorem]{Definition}

\DeclareMathOperator*{\argmax}{arg\,max}

\title{The Definitions of Interpretability and Learning of Interpretable Models}

\author{
   Weishen Pan\\ Tsinghua University
   \and
   Changshui Zhang \\ Tsinghua University
}
\date{May 2021}

\begin{document}

\maketitle

\begin{abstract}
As machine learning algorithms getting adopted in an ever-increasing number of applications, interpretation has emerged as a crucial desideratum. In this paper, we propose a mathematical definition for the human-interpretable model. In particular, we define interpretability between two information process systems. If a prediction model is interpretable by a human recognition system based on the above interpretability definition, the prediction model is defined as a completely human-interpretable model. We further design a practical framework to train a completely human-interpretable model by user interactions. Experiments on image datasets show the advantages of our proposed model in two aspects: 1) The completely human-interpretable model can provide an entire decision-making process that is human-understandable; 2) The completely human-interpretable model is more robust against adversarial attacks. 
\end{abstract}

\section{Introduction}

Interpretability has been considered an important property for machine learning models. It is becoming more important when the models are implemented as deep neural networks (DNNs). Most of them are end-to-end models that go directly from raw input $x$ (e.g., pixels) to target $y$ (e.g. object category). Such an issue will become more severe when a model provides a wrong prediction or is under adversarial attacks \cite{goodfellow2014explaining, madry2017towards, carlini2017towards}.

\begin{figure}[tb]
	\center{
		\includegraphics[width=0.5\columnwidth]{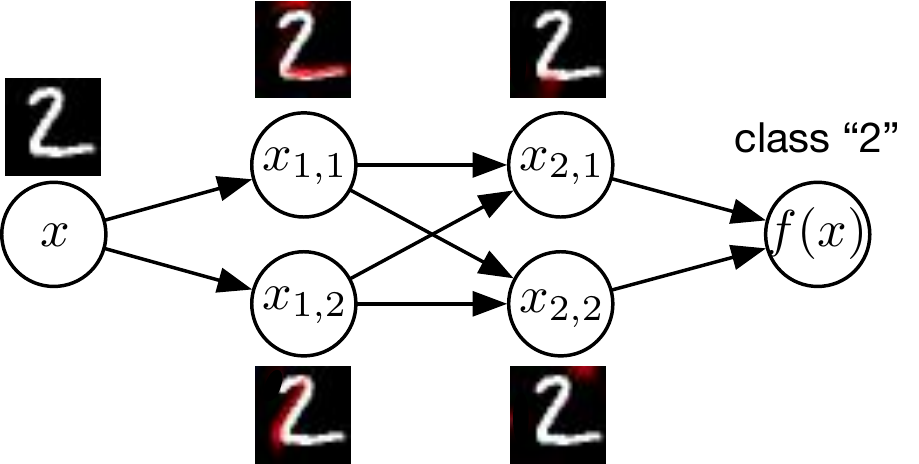}}
	\caption{The prediction process of a human-interpretable model $f$ on a given example $x$. $x_{1,1},x_{1,2},x_{2,1},x_{2,2}$ are feature maps of different neurons in the network, indicated by the red masks on original images. Best viewed in color. }
	\label{fig:intro}
\end{figure}

Interpretability is defined as “the ability to explain or to present in understandable terms to a human”. Although a formal definition of interpretation remains elusive, the overall goal is to obtain and transform information from models or their behaviors into a domain that humans can make sense of. We illustrate an example in Figure \ref{fig:intro}: when a model $f$ classifies a image of written number $x$, it first extract low-level features ($x_{1,1}$ and $x_{1,2}$, the values of feature maps are represented by red color) as edges with different directions; then it combine low-level features to get high-level features ($x_{2,1}$ and $x_{2,2}$, representing corner of "$\angle$" and "$\supset$" respectively) and use the high-level features to make the final decision $f(x)$. Obviously, $f$ is interpretable by humans on the given image since the prediction process is matched with one of the recognition process on the same image by humans. However, existing works have focused on providing post-hoc interpretation analysis on a trained model \cite{bau2017network, dalvi2019one, olah2020zoom} or trying to learn human-specified concepts in only one of the intermediate layers \cite{de2018clinically, bucher2018semantic, chen2020concept, koh2020concept}. So they do not guarantee to learn a model as interpretable as that in Figure \ref{fig:intro}.

In this paper, we focus on the definition of the human-interpretable prediction models and how to train such models. First, we propose a mathematical definition ty between two information processing systems. Since machine learning models and human cognitive systems are information processing systems, we define completely human-interpretable prediction models by extending the above interpretability definition. Second, we propose a learning framework to train a completely human-interpretable model with the help of human interactions. In particular, we train the model layer by layer and query the user to select the interpretable features iteratively. We also utilize different regularizations to boost the model to learn interpretable features. And our learning framework is compatible with the scenario that we can save the interpretable patterns identified in some datasets and then reuse them on a new dataset.

With the completely human-interpretable model trained by our framework, we can generate and visualize the inference process from input to final prediction for any given example. All features involved in this process are human-understandable. We can further tell how these features connect and affect each other. Another intuitive advantage of the completely human-interpretable model is robustness against adversarial attacks. Some previous works found that adversarial robustness and interpretability of a network are associated and can be improved simultaneously \cite{ross2018improving,boopathy2020proper}. Since completely human-interpretable models have higher interpretability, they should more robust against adversarial attacks. Such property is proven empirically on both semi-synthetic and real datasets.

The main contributions of this work is as follows:

\begin{enumerate}
	\item We systematically study the problem of interpretation and provide mathematical definitions of interpretability between two information processing systems. 
	\item We extend the above definitions to propose a novel concept of the completely human-interpretable model and design a practical learning framework to build such a model from data by human interactions.
	\item We show the advantages of our proposed completely human-interpretable model: 1) more human-understandable prediction process; 2) stronger robustness against adversarial attacks.
\end{enumerate}

\section{Related work}\label{sec:related}

\subsection{Post-hoc Interpretation Analysis}\label{sec:post-hoc}
Many post-hoc methods have been developed to interpret trained models, including recent works on using human-specified concepts to generate explanations \cite{bau2017network,kim2018interpretability,zhou2018interpretable,ghorbani2019towards}.
These techniques rely on models automatically learning those concepts despite not having explicit knowledge of them, and can be particularly useful when paired with models that attempt to learn more interpretable representations \cite{bengio2013representation,chen2016infogan,higgins2016beta,melis2018towards}. However, post-hoc methods will fail when the models do not learn these concepts, existing works show that the neurons in convolutional neural networks (CNN) trained without any interpretation constraints learn patterns without any semantic or perceptual meaning \cite{mu2020compositional,olah2020zoom}. These works can help us analyze whether features learned by a model are interpretable or not and find out what semantic concepts the interpretable features correspond to. However, they can not guarantee to learn interpretable models because they do not change the training process of models. While we instead directly guide models to be interpretable at training time.

\subsection{Learning Interpretable Models} 
There are some previous works on learning interpretable models by including additional information during the training process. They propose to learn {\em Concept Bottleneck Models} on data with annotations of pre-specified concepts. Such models first predict the concepts, then use those predicted concepts to make a final prediction of the outcome. More recently, deep neural networks with concept bottlenecks have re-emerged as targeted tools for solving particular tasks, e.g., retinal disease diagnosis\cite{de2018clinically}, visual question-answering\cite{yi2018neural}, and content-based image retrieval\cite{bucher2018semantic}. Some works also explore learning concept-based models via auxiliary datasets \cite{losch2019interpretability, chen2020concept}. But concept bottleneck models heavily rely on the availability and quality of provided concepts. When some important concepts fail to be specified and annotated, the performance of concept bottleneck models may be poor. A few works try to study mathematical properties of interpretable features and propose methods to boost learning these features \cite{yeh2019completeness} without the supervision of concepts. But this method has only been evaluated on a simple synthetic shape dataset. These works motivate us to match each feature to an interpretable concept, but they only make one layer of a model to be interpretable and rely heavily on the quality and completeness of concept annotations. While we target at making the sequential decision-making process instead of only one layer of a model interpretable. 

\subsection{Interactive Concept Learning} 
Approaches to interactively learn a concept-based representation have been proposed, several of which produce interpretable concept-based representations. One of them firstly introduces end-user interactive concept learning through which users can generate labels for relevant concepts in large datasets to train concept classifiers, which can be done with transparent classifiers \cite{amershi2009overview}. 
Interactive topic models have also been proposed to obtain low-dimensional representations aligned with user intuition, they constrain whether words should or should not appear in the same topic by interactions \cite{hu2014interactive, lund2018labeled}. In the computer vision domain, one previous work proposed to learn mid-level features for image classification by jointly finding predictive hyper-planes, and learning a model to predict the nameability of those hyper-planes \cite{parikh2011interactively}. But this kind of works also focuses on learning an interpretable representation, while our proposed framework target at making entire prediction process interpretable.

\section{Methodology}

In this section, we first define the interpretability between two information processing systems in Section \ref{sec:interpretable}. Second, we propose the concept of the human-interpretable model in Section \ref{sec:concept}. Finally, we design a practical algorithm to learn a completely human-interpretable model from data in Section \ref{sec:algorithm}.

\subsection{Definition of Interpretability}\label{sec:interpretable}

Suppose there are two information processing systems $A$: $f^{(a)}(x)$ for $x \in \mathcal{X}^{(a)}$ and system $B$: $f^{(b)}(x)$ for $x \in \mathcal{X}^{(b)}$. Here $f^{(a)}(x)$ and $f^{(b)}(x)$ are the outputs of the systems respectively. 

When we consider the situation that both $A$ and $B$ are black-box systems and we can only observe their outputs. We define how well system $B$ is understandable by system $A$ on sample $x$ as follows:

\begin{definition}
	For a sample $x \in \mathcal{X}^{(b)}$ and a scalar $\delta > 0$, if $x \in \mathcal{X}^{(a)}$ and $\texttt{dist} (f^{(a)}(x), f^{(b)}(x)) < \delta$, then system $B$ is $\delta$-interpretable ($\delta$-understandable) by $A$ with respect to sample $x$.
\end{definition}

\noindent Here $\texttt{dist}$ is a distance function depending on the type of output. We can use $L_1$/$L_2$ distance for continuous output or cross entropy for soft-max probabilities of categorical output. $\delta$ is a parameter to restrict how close $f^{(a)}(x)$ and $f^{(b)}(x)$ should be. $\delta = 0$ means $f^{(a)}(x)$ and $f^{(b)}(x)$ should be exactly the same, which is not practical in real-world problems sometimes. In practice, we set $\delta$ to be a small value. Intuitively speaking, if the outputs of two system on $x$ are different, they can not understand each other. So we define that if the outputs of two systems ($A$ and $B$) are almost the same on sample $x$ (measured by $\delta$), we say that $A$ can understand $B$ on sample $x$, or $B$ is understandable by $A$ on sample $x$. We also require $x \in \mathcal{X}^{(a)}$ explicitly. 

Next, we extend this definition to the case where $A$ and $B$ are white-box systems whose intermediate results can also be observed. As a sample $x$ passes through $B$, the sequential results step by step can be observed as $\{x_{0}^{(b)}=x, x_{1}^{(b)}, \dots, x_{N_{B}}^{(b)} = f^{(b)}(x)\}$ where $N_B$ is number of observed results. This sequence represents the information flow of how system $B$ processes sample $x$. And corresponding sequential results are also available for $A$. The interpretability of systems with the sequential results is defined as follows:

\begin{definition} \label{def:white-inter}
	For a sample $x \in \mathcal{X}^{(b)}$ and a scalar $\delta > 0$, if $x \in \mathcal{X}^{(a)}$ and there exists a sequence of the observed results of system $A$: $\{x_{0}^{(a)}=x, x_{1}^{(a)}, \dots, x_{N_{B}}^{(a)} = f^{(a)}(x)\}$ that $\texttt{dist}(x_{i}^{(a)}, x_{i}^{(b)}) < \delta, \forall i \in \{1,\dots,N_B\}$, then system $B$ is $\delta$-interpretable($\delta$-understandable) by system $A$ with respect to sample $x$.
\end{definition}

\noindent Here $x_{i}^{(a)}$ and $x_{i}^{(b)}$ can be scalars, vectors, matrices or graphs, depending on the structures of systems. And $\texttt{dist}$ is the corresponding distance function. Note that such definition is asymmetric. When $B$ is $\delta$-interpretable by $A$ with respect to sample $x$, $A$ can be more complex than $B$ and the resulting $\{x_{0}^{(a)}=x, x_{1}^{(a)}, \dots, x_{N_{B}}^{(a)} = f^{(a)}(x)\}$ is a sub-sequence of the entire sequantial results of $A$. So there might exist an intermediate result of $x$ from system $A$ which can not be matched with any observed result from $B$.

To quantitatively measure how well $B$ is $\delta$-interpretable by $A$ on $\mathcal{X}^{(b)}$, we propose the following definition:

\begin{definition}
	For a scalar $\delta > 0$, suppose $S_{\delta}(A,B)$ is the subset of $\mathcal{X}^{(b)}$ containing the samples where $B$ is $\delta$-interpretable by $A$ and $\frac{|S_{\delta}(A,B)|}{|\mathcal{X}^{(b)}|} = u$, then $B$ is $\delta$-interpretable by $A$ with $u$ degree.
\end{definition}

For simplicity, {\em $B$ is $\delta$-interpretable by $A$ with $u$ degree} is written as {\em $B$ is $u-\delta$ interpretable by $A$} in the rest of this paper. When we apply this definition in real-world problems, we need to consider the trade-off between $u$ and $\delta$. For two given systems $A$ and $B$, when $\delta$ becomes smaller, the proportion of samples on which $B$ is $\delta$-interpretable by $A$ (measured by $u$) will also become smaller.

\subsection{Completely Human-Interpretable Model} \label{sec:concept}

Let's focus on a prediction problem to predict a target $y$ from input $x$ with a model $f$. After sample $x$ is fed into $f$, the model will provide a sequence of observed results $\{x_{0}=x, x_{1}, \dots, x_{N} = f(x)\}$ where $N$ is the length of the sequence. Considering $x_{i}$ is a vector with $K_i$ dimensions, $x_{i,j}$ is the $j$-th value of $x_{i}$. Such setting can cover most mainstream machine learning models including but not limited to various kinds of neural networks, decision trees, etc.

Based on the definitions above, we propose to define whether a prediction model is interpretable by humans. We can treat the human cognitive system as $A$ and a given prediction model $f$ as $B$ and require $B$ is interpretable by $A$ by Definition \ref{def:white-inter}. Ideally we require $B$ to be $1-0$ interpretable (at least $1-\delta$ interpretable) by $A$. We require that each observed result $x_{i}^{(b)}$ from $B$ can be interpretable by humans according to Definition \ref{def:white-inter}. And we call such a model {\em completely human-interpretable model}. To judge whether prediction model $B$ is a completely human-interpretable model, one way is to find out one possible human-decision process as $A$ that $x_{i}^{(b)}$ and $x_{i}^{(a)}$ are almost the same (controlled by $\delta$) on almost all the data samples (controlled by $u$). However, it is very difficult to find out such a human-decision process for any $x$. Since there can exist multiple cognitive processes for humans to recognize a given object. For example, when we recognize a writing number in an image to be "4", we may decide in different ways: (1) the number is constructed by two strokes of "$\angle$" and "1" while "1" cross over "$\angle$" from upside down; (2) the number is constructed by a stroke "{\em 1}" and a shape "+". And it is very challenging to represent those cognitive processes due to the following reasons:

\begin{enumerate}
	\item  The number of cognitive processes for humans to recognize a given object might be infinite. Sometimes we may even not recognize some cognitive processes. For example, we may direct tell a number to be "4" by intuition, without decomposing it into strokes explicitly. So it will be very hard for us to list all possible cognitive processes. 
	\item We are not able to describe the space to represent the infinite cognitive processes. So it is also unrealistic for us to find a human cognition process from this space by some existing techniques (such as calculating gradients). 
	\item We know little about the complex mechanisms of the human cognitive system, rather than represent them with mathematical formulas.
\end{enumerate}

In conclusion, for a given prediction model $B$, it is difficult for us to consider all possible human cognitive processes and judge whether $B$ is matched with one of them by Definition \ref{def:white-inter}. Due to the same reasons, it is also not practical for us to select and implement one of the human recognition systems as $A$ and generate a prediction model $B$ which is interpretable by $A$.

It is hard for us to formulate an entire human recognition system and compare it with model $B$. Such a system can be a hierarchical system consisting of semantic concepts organized by human cognitive logic. But it is possible to match one of the observed values $x_{i,j}$ to a meaningful concept \cite{koh2020concept,chen2020concept}, which can be lexical or perceptual. For example, in the computer vision domain, the meaningful concept can be red color, horizontal line, shape as vertical rails, or head of a dog. Since human annotations on pre-specified meaningful concepts are usually unavailable in real-world scenarios, we propose to solve this problem by interaction with humans. In particular, we display the values of $x_{i,j}$ on some data samples to a user and query the user about whether $x_{i,j}$ is associated with a meaningful concept. We conduct such a process to $x_{i,j}$ one by one until all observed results are interpretable by humans. The algorithm is shown in the next section.

\subsection{Algorithm} \label{sec:algorithm}

According to the discussion in Section \ref{sec:post-hoc}, existing machine learning models have the potential to learn semantic concepts. However, it is not guaranteed that all intermediate results of a model are interpretable. One intuitive solution is to train a model and ask a user to pick up all the interpretable features. But if we only conduct such a process in a single run, the model may fail to learn plenty of interpretable representation for the user to pick. To make sure interpretable and predictive patterns can be obtained as many as possible, we propose a framework to train a human-interpretable model iteratively. 

For the sake of simplicity, we use a multi-layer neural network as an example to illustrate the process of our algorithm. And our algorithm can be extended to other machine learning models which can provide sequential results. Suppose $f$ is a neural network with $L$ layers. For a sample $x$, $x_i$ is the output of $i$-th layer with $K_i$ neurons and $x_{i,j}$ is the $j$-th value of $x_{i}$. $\theta$ is the set of parameters of the entire model $f$, while $\theta_{i}$ is the set of parameters of the $i$-th layer.

We train a completely human-interpretable network layer by layer. When we train the $i$-th layer, we freeze the parameters of the previous $i-1$ layers. $\mathbf{S}_i$ is the set of neurons that have been judged by the user to be interpretable and initialized as $\mathbf{S}_i = \emptyset$. We use $x_{i, \mathbf{S}_i}$ and $\theta_{i, \mathbf{S}_i}$ to represent the values of neurons in $\mathbf{S}_i$ and corresponding parameters. We also denote the remaining neuron in the $i$-th layer as $\bar{\mathbf{S}}_i = \{1,\dots,K_i\} \setminus \mathbf{S}_i$. We train the models and select neurons into $\mathbf{S}_i$ iteratively. In the first iteration, we train the parameters from layer $i$ to $L$ to minimize an objective function consisting of the prediction loss. Then the extracted features $x_{i,j}$ on a sample set for each $j \in \bar{\mathbf{S}}_i$ are displayed to a user, and the user should choose the meaningful neurons and add them into $\mathbf{S}_i$. After a neuron is selected into $\mathbf{S}_i$, its parameters will be frozen. We repeat the iterations until $\mathbf{S}_i$ contains all neurons in $f_i$, or no more neurons in $\bar{\mathbf{S}}_i$ is judged to be interpretable in a certain iteration. To encourage the network to learn more distinguishable patterns, we add a sparsity regularization on the features in layer $i$ along with the original prediction loss. As the iterations go on, to explore new features which are different from those in $\mathbf{S}_i$, we add a regularization as the correlation among features in $\mathbf{S}_i$ and those in $\bar{\mathbf{S}}_i$. The function of prediction loss, sparsity and correlation regularizations are denoted as $\mathcal{L}_{pred}$, $\mathcal{L}_{s}$ and $\mathcal{L}_{c}$ respectively. In practice, we can implement $\mathcal{L}_{pred}$ as cross-entropy for classification and mean squared error for regression. For $\mathcal{L}_{s}$, we use the $L_1$ norm of the intermediate features:

\begin{equation}
\mathcal{L}_{s} = \frac{1}{M |\bar{\mathbf{S}}_i|}\sum_{x} \left\|x_i\right\|_1,
\end{equation}
\noindent where $M$ is the number of samples.

$\mathcal{L}_{c}$ is defined to be 0 when $\mathcal{S}_i \neq \emptyset$ and as follows otherwise:

\begin{equation}
\label{equ:loss}
\mathcal{L}_{c} = \frac{1}{|\bar{\mathbf{S}}_i||\mathbf{S}_i|}\sum_{j \in \mathbf{S}_i} \sum_{\bar{j} \in \bar{\mathbf{S}}_i}| \texttt{Corr}(\mathbf{x}_{i,j}, \mathbf{x}_{i,\bar{j}})|,
\end{equation}

\noindent where $\texttt{Corr}$ is the correlation function and $\mathbf{x}_{i,j}$ is the vector obtained by aggregating all $x_{i,j}$ of the dataset.

The weighted loss function we optimize during training is:

\begin{equation}
\mathcal{L} = \mathcal{L}_{pred} + \lambda_{s} \mathcal{L}_s + \lambda_{c} \mathcal{L}_c,
\end{equation}

\noindent where $\lambda_{s}$ and $\lambda_{c}$ are parameters. The overall process is shown in Algorithm \ref{alg:algorithm}. Note that when we train the $i$-th layer, $\theta_{i+1:L}$ will also be optimized. Such process is to make sure the features $x_{i}$ are predictive of the outcome.

\begin{algorithm}
	\begin{algorithmic}[1]
		\STATE{Input: $\{(x, y)\}$}
		\STATE{Output: a completely human-interpretable model $f$}
		\STATE{Initialize: $f$}
		\FOR{$i$ in 1:$L-1$}
		\STATE{Initialize: $\mathbf{S}_i = \emptyset$}
		\WHILE{True}
		\STATE{Optimize $\mathcal{L}$ in Eq. (\ref{equ:loss}): $\argmax \limits_{\theta_{l,\bar{\mathbf{S}}_i}, \theta_{l+1:L}} \mathcal{L}$}
		\STATE{}
		\FOR{$j$ in $\bar{\mathbf{S}}_i$}
		\IF{$x_{i,j}$ is user-interpretable}
		\STATE{Add $j$ to $\mathbf{S}_i$}
		\ENDIF
		\ENDFOR
		\IF{$|\bar{\mathbf{S}}_i| = 0$ or $\mathbf{S}_i$ does not grow}
		\STATE{break}
		\ENDIF
		\ENDWHILE
		\ENDFOR
	\end{algorithmic}
	\caption{Algorithm to train a completely human-interpretable model}\label{alg:algorithm}
\end{algorithm}

\subsection{Storage and Reusage of Interpretable Patterns}

Human-interpretable concepts should be shared across different datasets. For example, colors and textures are shared important patterns in almost all image datasets. This motivates us to design a pipeline to store the semantic concepts identified by the user after training completely human-interpretable models on some datasets and reuse them when we encounter a new dataset. 

First, we consider how to store and manage the semantic concepts. We propose to maintain a concept pool $\mathbf{C}$. After we train a completely human-interpretable model $f^{*}$ on a dataset $\{(x^{*}, y^{*})\}$ consisting $M^{*}$ samples, suppose $x^{*}_{i,j}$ is identified to be corresponding to a concept $\mathbf{c}$. One way is to directly store $f^{*}$ and the index of $(i,j)$. However, it would be hard to manage such a concept pool. Since the structures of $f^{*}$ on different datasets may be different. Moreover, it is redundant to save the whole model $f^{*}$ when we only want to save some of the patterns. So we re-train a concept detector $g_{\mathbf{c}}$ by minimizing that following objective function: $\frac{1}{M^{*}}\sum_{x^{*}} \texttt{dist}(g_{\mathbf{c}}(x^{*}), x^{*}_{i,j})$. Then $\mathbf{c}$ would be added to $\mathbf{C}$ and $g_{\mathbf{c}}$ will also be saved. 

We extend Algorithm \ref{alg:algorithm} into the form when we obtain a concept pool $\mathbf{C}$ and corresponding detectors $\{g_{\mathbf{c}}\}$. The mainstream of the algorithm is almost the same. The difference is that before the user interact with features in $\bar{\mathbf{S}}_i$ to decide which are meaningful, we first try to match them with the concepts in $\mathbf{C}$. According to the definitions in Section \ref{sec:interpretable}, for given $g_{\mathbf{c}}$, if $\frac{\sum_{x}\mathcal{I}(\texttt{dist}(g_{\mathbf{c}}(x), x_{i,j}) < \delta)}{M} > u$ where $u,\delta$ are given parameters, then we can say the concept $\mathbf{c}$ can represent $x_{i,j}$. The proposed algorithm is conducted by inserting in Module \ref{alg:algorithm_concept} into Line 8 of Algorithm \ref{alg:algorithm}, where $\mathcal{I}$ is the indication function, $\delta$ and $u$ are parameters to determine how close $g_{\mathbf{c}}(x)$ and $x_{i,j}$ should be.

\setcounter{algorithm}{0}
\floatname{algorithm}{Module}
\begin{algorithm}
	\begin{algorithmic}[1]
		\STATE{Input: $\{(x, y)\}$, $\{g_{\mathbf{c}}\}, \mathbf{c} \in \mathbf{C}$, $\delta$, $u$}
		\FOR{$\mathbf{c}$ in $\mathbf{C}$}
		\IF{$\frac{\sum_{x}\mathcal{I}(\texttt{dist}(g_{\mathbf{c}}(x), x_{i,j}) < \delta)}{M} > u$}
		\STATE{Add $j$ to $\mathbf{S}_i$}
		\STATE{break}
		\ENDIF
		\ENDFOR
	\end{algorithmic}
	\caption{Additional module to train a completely human-interpretable model with pre-identified concepts}\label{alg:algorithm_concept}
\end{algorithm}

\section{Interpretation and Robustness}

Robustness against adversarial attacks is another important property for deep neural networks \cite{goodfellow2014explaining}. The data instances after being attacked are called adversarial samples. Clean samples and corresponding adversarial samples are usually not distinguishable to humans, while adversarial samples may be predicted dramatically differently from clean samples by a machine learning model. Previous works have found that robustness ad and interpretability of models are associated. For example, the representations learned by robust models tend to align better with salient data characteristics and human perception \cite{tsipras2018robustness}. From another perspective, a well-interpretable model may be more robust. We will explain it in the next paragraph.

The vulnerability of a prediction model under adversarial attacks comes from two reasons. One includes the number of samples and learning algorithms, which can be alleviated by training with more samples or improving training algorithms \cite{goodfellow2014explaining,madry2017towards,lu2017safetynet}. The second reason is the model itself. The mechanism of a prediction model implemented on computers is different from that of a human cognitive system. So it is reasonable that the model will make wrong decisions on adversarial examples while humans will not. People will think the model is not robust from the view of humans. However, this phenomenon may be normal from the view of the model itself. If we take a deeper look, robustness against adversarial attacks can be seen as the byproduct of interpretability. Since humans can correctly identify the interpretable features and make decisions upon these features, the human cognitive system is robust against adversarial attacks. For example, while humans classify a writing number, they identify different kinds of strokes by their directions and make the final decision based on the positions of strokes. Such characteristics can hardly be changed by slight changes on some pixels. However, a prediction model is implemented to recover the input-output relation from data, without considering interpretability explicitly. So the model is not guaranteed to be robust against the human-indistinguishable permutations. And we believe designing a model which is interpretable by humans can alleviate or even eliminate the impact of the second reason. 

\section{Experiments}

We evaluate the advantages of the completely human-interpretable model learned by our framework in the following aspects: 1) With the help of human interactions, we can train a completely human-interpretable model which can provide a decision process in which each component is aligned with human-understandable concepts. 2) We examine the robustness of models with two adversarial attack methods and show that our proposed model is more robust than that trained without any interpretation constraints.

\subsection{Datasets}

\noindent \textbf{MNIST} A benchmark dataset containing images of writing letters from 0 to 9. It consists of stroke patterns such as lines and curves. Although MNIST has been widely used to analyze the behaviors of CNNs, no annotations on intermediate patterns are available.

\noindent \textbf{CMNIST} A semi-synthetic dataset generated from the MNIST dataset, in which patterns as colors and lines. We sample the examples of the MNIST dataset of digits 1, 4, 7 and randomly paint them with red, green, and blue. A combination of possible color and digit corresponds to one category of object. And we can assume the ground-truth human-interpretable patterns in this dataset to be colors of red/green/blue, horizontal/vertical lines.

\subsection{Implementation Details}

\noindent \textbf{Network Structure.} The networks structures used in the datasets are shown in Tables \ref{tab:cmnist_arch} and \ref{tab:mnist_arch} respectively. We treat a block of Conv, Relu, and max-pooling as a layer. $x_{i}$ is the output from the $i$-th layer,  $f_{i,j}$ is the $j$-th channel of the $i$-th layer. $f_{i,j}(x)$ is the corresponding feature map of sample $x$. For CMNIST, we implement the network structure as in Table \ref{tab:cmnist_arch} with only one intermediate layer since the patterns involved is simple. For MNIST, we use two intermediate layers to learn different kinds of strokes.

\noindent \textbf{Training Process.} We use the original train/test split as in the MNIST dataset. And when we generate the data in CMNIST from MNIST, we follow the train/test split. All models are optimized with the ADAM algorithm. The learning rate is set to be $0.001$. And other parameters are set as: $\lambda_s = 0.8$ and $\lambda_c = 0.1$.

When we train a completely human-interpretable network on the CMNIST dataset, we save the interpretable patterns and reuse them on the MNIST dataset.

\begin{table}[h]
	\centering
	\caption{Network architecture of CMNIST dataset.}
	\label{tab:cmnist_arch}
	\begin{tabular}{|l|l|}
		\hline
		Meaning & Structure \\ \hline
		$x$ & Input 28 $\times$ 28 $\times$ $3$ image \\ \hline
		\multirow{2}{*}{$x_1$} & 5 $\times$ 5 conv, 5 channels, stride 2; ReLU \\
		& 2 $\times$ 2 maxpooling.\\ \hline
		$f(x)$ & FC 9, softmax\\ \hline
	\end{tabular}
\end{table}

\begin{table}[h]
	\caption{Network architecture of MNIST dataset.}
	\label{tab:mnist_arch}
	\centering
	\begin{tabular}{|l|l|}
		\hline
		Meaning & Structure \\ \hline
		$x$ & Input 28 $\times$ 28 $\times$ $1$ image \\ \hline
		\multirow{2}{*}{$x_1$} & 5 $\times$ 5 conv, 10 channels, stride 1; ReLU \\ 
		& 2 $\times$ 2 maxpooling.\\ \hline
		\multirow{2}{*}{$x_2$} & 5 $\times$ 5 conv, 20 channels, stride 1; ReLU \\
		& 2 $\times$ 2 maxpooling.\\ \hline
		$f(x)$ & FC 10, softmax\\ \hline
	\end{tabular}
\end{table}


\subsection{Interpretable Process when Model Makes Predictions}

\begin{figure}[tb]
	\center{
		\includegraphics[width=1.0\columnwidth]{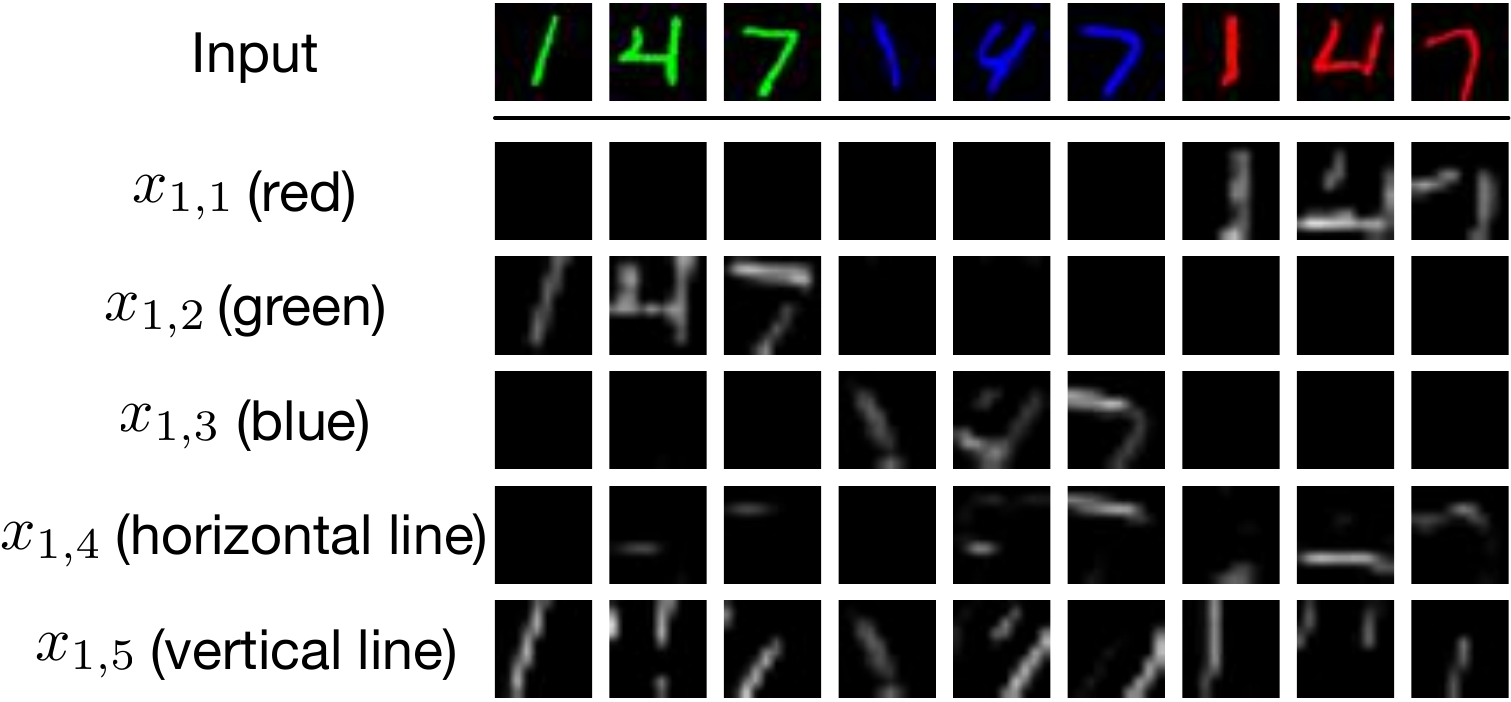}}
	\caption{The visualization result of human-interpretable neurons learned by our proposed method. Each row represents the feature maps of each neuron on a set of samples. Best viewed in color.}
	\label{fig:visual_cmnist}
\end{figure}

\begin{figure}[tb]
	\center{
		\includegraphics[width=1.0\columnwidth]{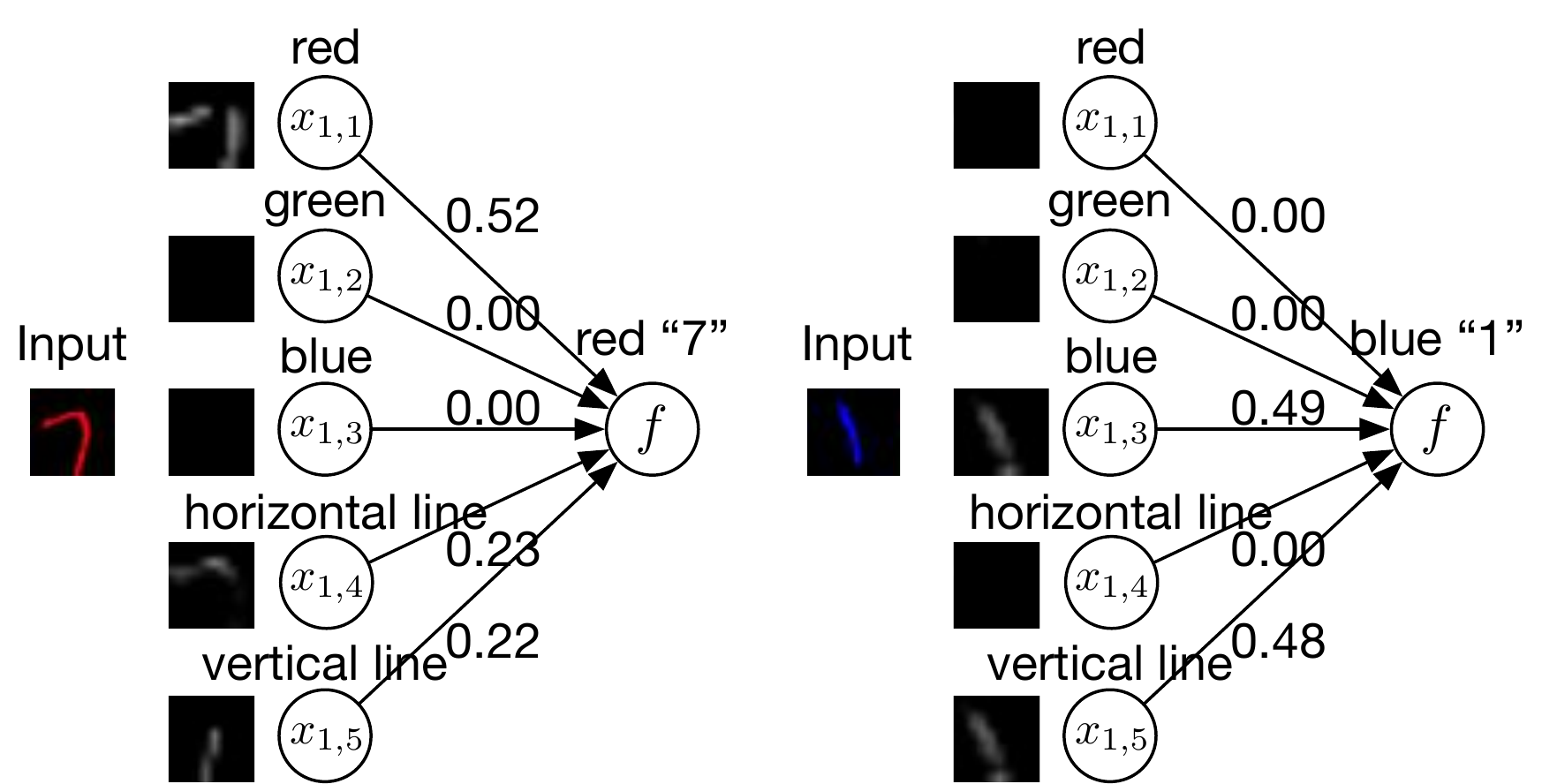}}
	\caption{The sequential process of how the model correctly classifies an image of red "7" and an image of blue "1". The contributions of each neuron to the predicted probability of the groud-truth label are shown. Best viewed in color.}
	\label{fig:process_cmnist}
\end{figure}

We show the completely human-interpretable network learned by our algorithm on the CMNIST dataset in Figure \ref{fig:visual_cmnist}. We present the feature maps of each neuron on a set of samples to qualitatively show what concept is learned by each neuron. We can see that each neuron in the network learned by our algorithm is matched with one ground-truth human-interpretable pattern. With a completely human-interpretable model, we can obtain its decision-making process on any given data sample. We present the examples in Figure \ref{fig:process_cmnist}. When the model classifies an image of blue "1", it extracts the intermediate features as the blue color and vertical strokes from the image. Then it combines these two patterns to come out with the final decision.

\begin{figure}[tb]
	\center{
		\includegraphics[width=1.0\columnwidth]{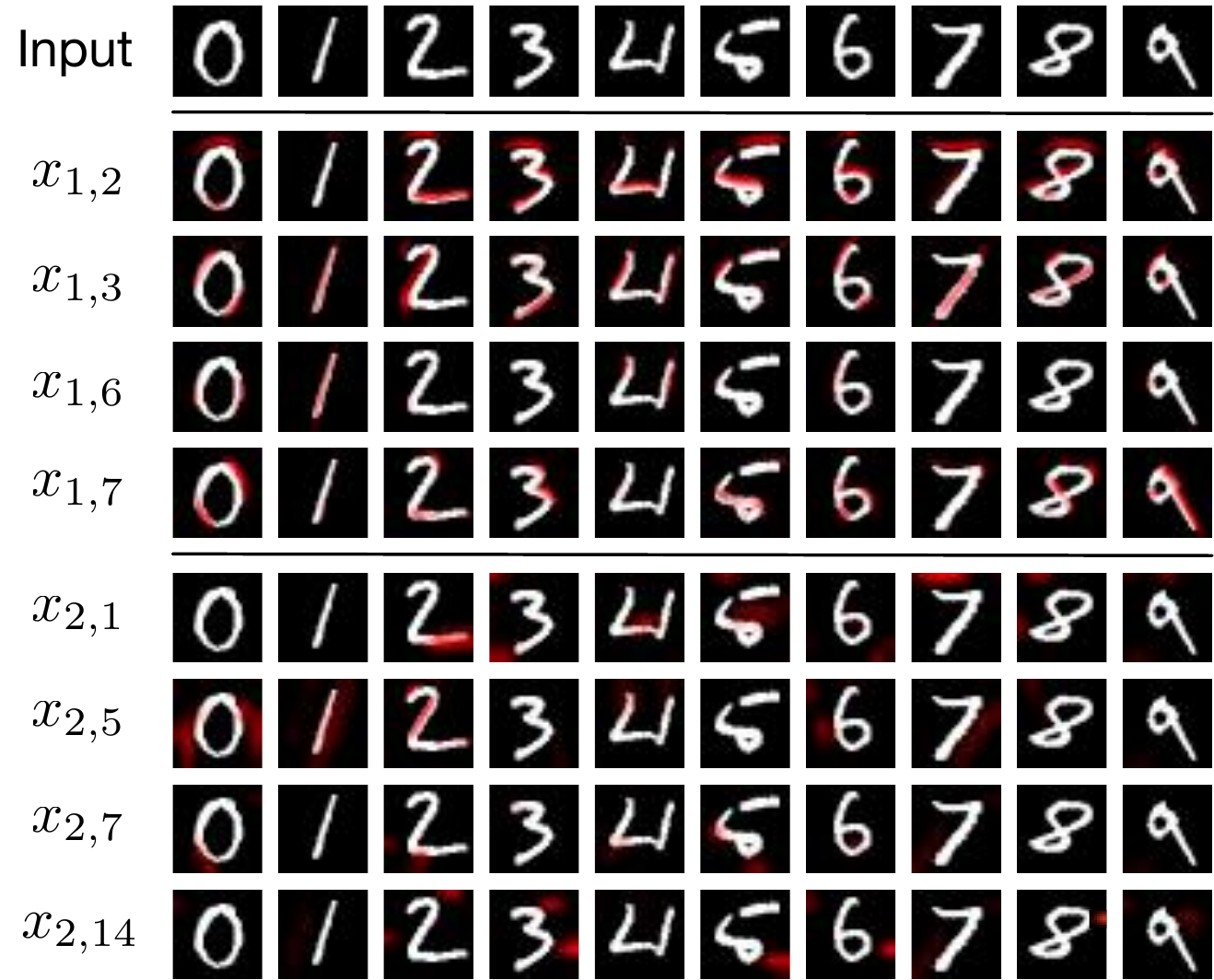}}
	\caption{The visualization result of human-interpretable neurons was learned by our proposed method on the MNIST dataset. Each row represents the feature maps of each neuron on a set of samples. For clear representation, we display the original image with white strokes and mask it with a red feature map.}
	\label{fig:visual_mnist}
\end{figure}

\begin{figure}[tb]
	\center{
		\includegraphics[width=1.0\columnwidth]{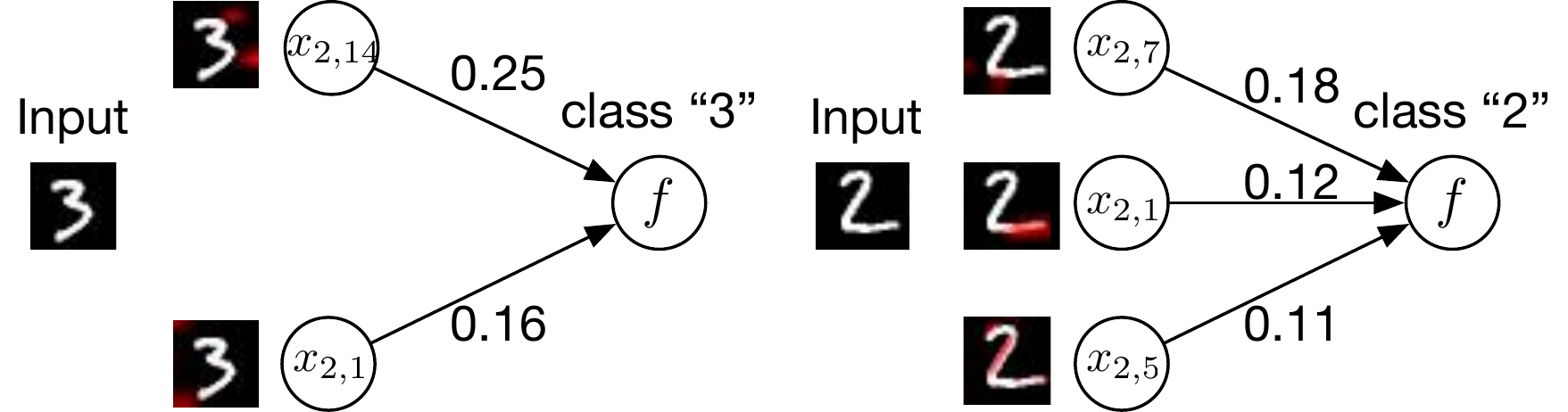}}
	\caption{The sequential process of how the model correctly classifies an image of class "3" and an image of class "2". The contributions of each neuron to the predicted probability of the groud-truth label (by connection weight) are shown. Best viewed in color. }
	\label{fig:process_mnist}
\end{figure}

The results on MNIST are in Figure \ref{fig:visual_mnist} and \ref{fig:process_mnist}. We selectively display feature maps for some neurons to show what our proposed model has learned. We can see that the first layer learns local patterns representing different directions of lines. For example, $x_{1,2}$ represents the horizontal line while $x_{1,7}$ represents the line tilted 45 degrees. Such line patterns can also be observed in the second layer, along with more complex patterns. For example, $x_{2,7}$ learns the corner of "$\angle$", which occurs in "2", "4" and "5". And $x_{2,14}$ learns the "$\supset$", which occurs in "2", "3", "5", "6", "8" and "9". In Figure \ref{fig:process_mnist}, we present two examples of how the model classifies images of class "2" and "3" as the ground-truth labels. We show the features which contribute most to the predicted probability of the ground-truth label. The features used to predict are matched with human recognition: if an image contains "$\angle$" ($x_{2,7}$) on the bottom-left, horizontal line ($x_{2,1}$) on the bottom and tilted line ($x_{2,5}$) in the middle, then the image is classified to be number "2".

\subsection{Robustness under Adversarial Attacks}

\noindent \textbf{Adversarial Attack Methods.} First, to evaluate the robustness of classification, we consider conventional PGD attacks with different steps and perturbation sizes \cite{madry2017towards} and C$\&$W-type attack \cite{carlini2017towards}. Unless specified otherwise, we choose the perturbation size $\epsilon=0.2$ on both datasets.

\begin{table}
	\caption{Accuracy under PGD adversarial attack on both datasets under different levels of permutations.}
	\label{tab:comparison_pgd}
	\centering
	\begin{tabular}{|c|c|cccc|}
		\hline
		Dataset & Model & $\epsilon$ = 0 & 0.1 & 0.2 & 0.3\\ \hline
		\multirow{3}{*}{CMNIST} & Baseline & 0.985 & 0.088 & 0.033 & 0.020\\
		& Sparse & 0.985 & 0.190 & 0.035 & 0.025\\ 
		& Ours & 0.986 & {\bf 0.321} & 0.037 & 0.022\\ \hline
		\multirow{3}{*}{MNIST} & Baseline & 0.984 & 0.712 & 0.265 & 0.037\\
		& Sparse & 0.984 & 0.777 & 0.355 & 0.051\\
		& Ours & 0.984 & \bf{0.791} & \bf{0.392} & \bf{0.061}\\ \hline
	\end{tabular}
\end{table}

\begin{table}
	\caption{Accuracy under C$\&$W adversarial attack on both datasets under different levels of permutations. }
	\label{tab:comparison_cw}
	\centering
	\footnotesize
	\begin{tabular}{|c|c|ccccc|}
		\hline
		Dataset & Model & $\epsilon$ = 0 & 0.1 & 0.2 & 0.3 & 0.4\\ \hline
		\multirow{3}{*}{CMNIST} & Baseline & 0.985 & 0.746 & 0.534 & 0.357 & 0.254\\
		& Sparse & 0.985 & 0.867 & 0.705 & 0.516 & 0.384\\ 
		& ours & 0.986 & {\bf 0.882} & {\bf 0.740} & {\bf 0.579} & {\bf 0.469}\\ \hline
		\multirow{3}{*}{MNIST} & Baseline & 0.984 & 0.880 & 0.658 & 0.396 & 0.192\\
		& Sparse & 0.984 & 0.910 & 0.765 & 0.481 & 0.234\\
		& ours & 0.984 & 0.911 & \bf{0.785} & \bf{0.503} & \bf{0.274}\\ \hline
	\end{tabular}
\end{table}

\noindent \textbf{Baselines.} Since there is no existing method proposed to a completely human-interpretable model like ours. We choose the network with the same network structure but trained without any sparsity or interpretability constraints as one of the baselines (called {\em Baseline}). To further evaluate the advantage of our model that comes from interpretability rather than sparsity, we also train a baseline model with sparsity regularization (called {\em Sparse}).

Table \ref{tab:comparison_pgd} and \ref{tab:comparison_cw} report the classification accuracy under adversarial attacks over CMNIST and MNIST. In the tables, each row represents the accuracy of the model under different levels of attack. $\epsilon$ indicates the level of attack, while $\epsilon = 0$ means accuracy under clear samples. We can see that our proposed model is more robust than baselines, while it achieves almost the same performance on clean images. And the robustness should be beneficial from the interpretability of the model since our model outperforms the model with only sparsity regularization.

\section{Conclusion}

In this paper, we propose a novel concept of human- interpretable model and a practical framework to learn such a model from data. We empirically show its advantages as providing an interpretable prediction process and being robust against adversarial attacks. Moreover, our definition and learning framework is general and can be applied to various kinds of prediction tasks and models. And it is easy to extend such a framework to improve the quality of interpretable features. As future works, we can investigate designing regularizations or applying domain knowledge.

\bibliographystyle{apalike}
\bibliography{main}

\end{document}